\documentclass{INTERSPEECH2023}

\interspeechcameraready 
\usepackage{multirow}
\usepackage{diagbox}
\usepackage{hyperref}
\hypersetup{
    colorlinks=true,
    linkcolor=blue,
    filecolor=magenta,      
    urlcolor=cyan
}
\title{Score-balanced Loss for Multi-aspect Pronunciation Assessment}
\name{Heejin Do$^1$, Yunsu Kim$^{1,2}$, Gary Geunbae Lee$^{1,2}$}
\address{
  $^{1}$Graduate School of AI, POSTECH, Republic of Korea\\
  $^2$Department of Computer Science and Engineering, POSTECH, Republic of Korea}
\email{\{heejindo, yunsu.kim, gblee\}@postech.ac.kr}

\begin{document}

\maketitle
 
\begin{abstract}
With rapid technological growth, automatic pronunciation assessment has transitioned toward systems that evaluate pronunciation in various aspects, such as fluency and stress. However, despite the highly imbalanced score labels within each aspect, existing studies have rarely tackled the data imbalance problem. In this paper, we suggest a novel loss function, score-balanced loss, to address the problem caused by uneven data, such as bias toward the majority scores. As a re-weighting approach, we assign higher costs when the predicted score is of the minority class, thus, guiding the model to gain positive feedback for sparse score prediction. Specifically, we design two weighting factors by leveraging the concept of an effective number of samples and using the ranks of scores. We evaluate our method on the speechocean762 dataset, which has noticeably imbalanced scores for several aspects. Improved results particularly on such uneven aspects prove the effectiveness of our method.
\end{abstract}
\noindent\textbf{Index Terms}: automated pronunciation assessment, imbalanced dataset, score-balanced loss

\section{Introduction}
Automatic pronunciation assessment supports non-native (L2) language learners in acquiring foreign spoken languages as part of a computer-assisted pronunciation training (CAPT) system \cite{eskenazi2009overview, franco1997automatic}. With the advantages of immediate feedback and convenience, CAPT has been actively studied mainly in assessing pronunciation with a phoneme-level score \cite{witt2000phone, shi2020context, sancinetti2022transfer}. 

With increasing demands for detailed feedback, recent studies have evaluated pronunciation in several aspects of various granularity levels, such as stress, fluency, and prosody \cite{tepperman2005automatic, cucchiarini2000quantitative, arias2010automatic, li2017intonation}. As an alternative to multiple models that separately assess different aspects, current joint models \cite{gong2022transformer, chao20223m, 10095733} facilitate simultaneous prediction of multiple aspects using a single model considering the association between aspects.

However, notwithstanding the technical advances, extremely imbalanced score labels within each aspect that show high-score-biased distributions have rarely been studied on the pronunciation assessment task. Biased datasets of the CAPT system have been frequently reported \cite{yang2014machine, sancinetti2022transfer, gong2022transformer, chao20223m}; however, they are rarely optimized for the scoring task. Imbalanced datasets could cause the model to be overfitted toward the majority classes during the training \cite{wang2017learning, cui2019class}; therefore, addressing the imbalance problem is crucial for qualified assessment. Furthermore, in the case of multi-aspect pronunciation scoring, uneven data distributions within certain aspects cause huge performance gaps between different aspects, which consequently hinders application to real-world educational situations. For practical use, accurately evaluating every aspect is important, without bias toward a specific aspect; in particular, good overall quality is better than a part with exceptional quality. 

In this paper, we introduce a novel loss function, score-balanced (SB) loss, to overcome the quality degradation caused by imbalanced data in multi-aspect pronunciation assessment. Motivated by the class-balanced loss suggested for visual classification \cite{cui2019class}, we introduce cost-sensitive re-weighting schemes for balancing the weight when scoring pronunciation. Unlike class-balanced loss, which is based on the ground-truth label class, the proposed $SB_{num}$ loss directly targets the predicted scores and aims at score-labeled regression. Specifically, we design SB loss with two different factors: $SB_{num}$, which leverages the number of samples on categorized scores, and $SB_{rank}$, which exploits those ranks. By assigning high costs to the predicted score of the minority class, the model is directed to favorably predict sparse classes despite fewer training samples.

We evaluate the proposed loss function on the public speechocean762 dataset, which is widely used for multi-aspect pronunciation assessments \cite{zhang2021speechocean762}. We find highly imbalanced data distributions, which are densely distributed toward a high score, on certain aspects such as \textit{Completeness} and \textit{Stress}. Attributing this result to the particularly low assessment qualities on such aspects in existing studies \cite{gong2022transformer, chao20223m, 10095733}, we train the open-source model, goodness of pronunciation feature-based transformer (GOPT) \cite{gong2022transformer}, with the proposed SB loss. Significantly improved results on the notably imbalanced aspects prove that the issue of unevenly distributed datasets has been successfully resolved, reducing the gap between different aspect assessment qualities. It is noteworthy that our enhancements are achieved without any augmentation or architecture modeling. Our codes are available on the GitHub link\footnote{\url{https://github.com/doheejin/SB_loss_PA}}.

\section{Related work}
A dataset is an essential factor for both supervised classification and regression. Therefore, the data imbalance problem has been actively discussed for decades, particularly around visual and text classification tasks \cite{he2009learning, japkowicz2002class, wang2017learning, padurariu2019dealing}. Generally, the related studies are divided according to two main approaches: re-sampling and re-weighting. Re-sampling over-samples or under-samples the data by repeating or deleting the existing data examples \cite{chawla2002smote, estabrooks2004multiple}, while re-weighting adjusts the loss function by assigning more costs to under-represented-class samples \cite{wang2017learning, cui2019class}. Recently, pointing out the lack of research on imbalanced regression which handles continuous targets, few studies dealing with non-categorical imbalanced data have been proposed, mostly for the visual dataset \cite{ren2022balanced, yang2021delving}. Emphasizing the need for a balancing strategy for challenging imbalanced regression, particularly for speech pronunciation assessment, we propose a novel loss optimized for score-labeled datasets.

Real-word assessments for educational use have suffered from naturally imbalanced datasets \cite{basuki2018use}, which tend to have biased grades. To improve the pronunciation error detection of CAPT by addressing problems caused by extremely imbalanced class datasets, machine learning-based re-weighting methods have been investigated \cite{yang2014machine}. Most recently, to compensate for the score imbalance in pronunciation assessment datasets, a transfer learning-based study \cite{sancinetti2022transfer} applied simple balanced weights with the inverse of the number of frames. Considering the score labels that are biased toward high values, GOPT \cite{gong2022transformer} set the Pearson correlation coefficient (PCC) as the major evaluation metric. However, their focus is not on data inequality, and an intensive investigation for tackling score unbalance is lacking. We propose a novel solution focusing on scoring data imbalance in multi-aspect pronunciation assessment.

%To the best of our knowledge, the present study is the first to propose a solution focused on scoring data imbalance in multi-aspect pronunciation assessment. 

\section{Methods}

\begin{figure}[t]
    \centering \includegraphics[width=8cm]{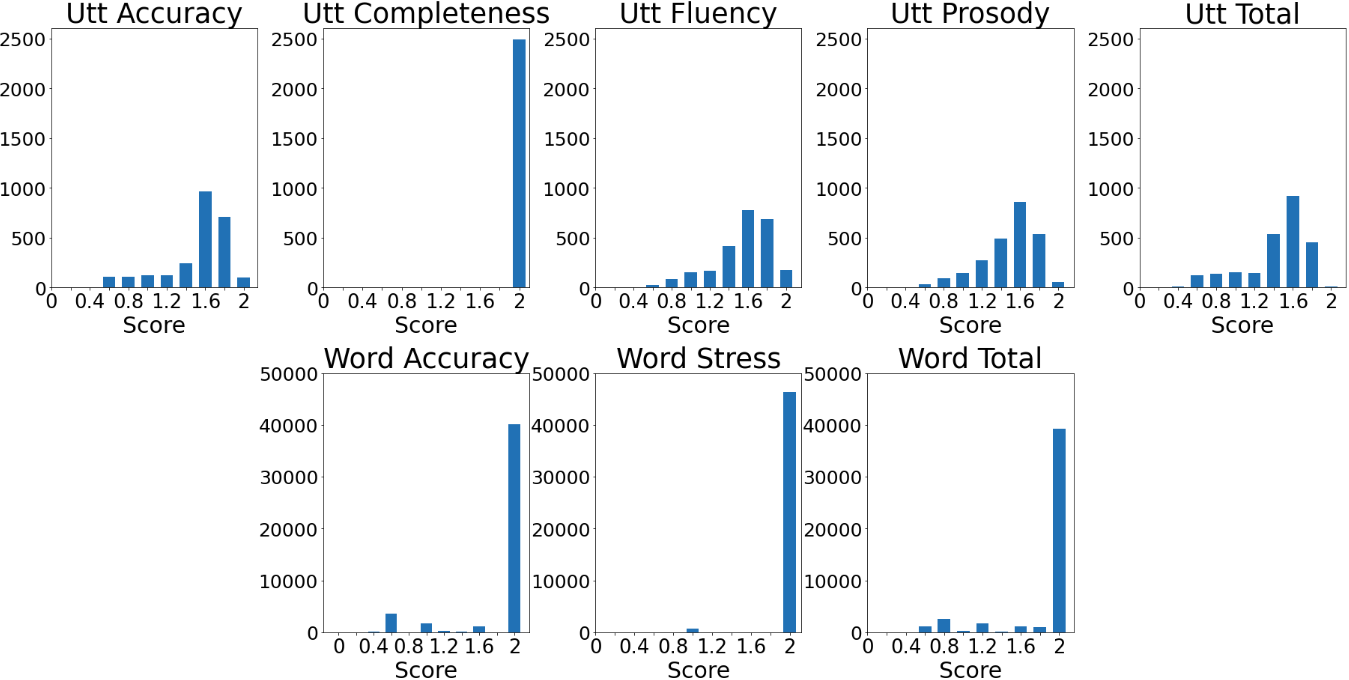}
    \caption{
Data distribution histogram across score labels for each aspect of the utterance (1st row) and word (2nd row).}
    \label{fig1}
\end{figure}

\subsection{Score-label imbalance dataset}
We investigate the score label distribution of the pronunciation assessment data by each aspect before presenting our method. We observe the publicly available speechocean762 dataset \cite{zhang2021speechocean762}, which provides rich score labels for various aspects and has been mainly used for multi-aspect pronunciation assessment tasks \cite{gong2022transformer, chao20223m, 10095733}. For each non-native speaker utterance, the dataset comprises phoneme-level \textit{Accuracy} score; word-level \textit{Accuracy}, \textit{Stress}, and \textit{Total} scores; and utterance-level \textit{Accuracy}, \textit{Completeness}, \textit{Fluency}, \textit{Prosody}, and \textit{Total} scores. The phoneme-level aspect has a score between 0-2, and the word-level and utterance-level aspects have a score between 0-10.

\begin{figure}[t]
    \centering
    \includegraphics[width=7.8cm]{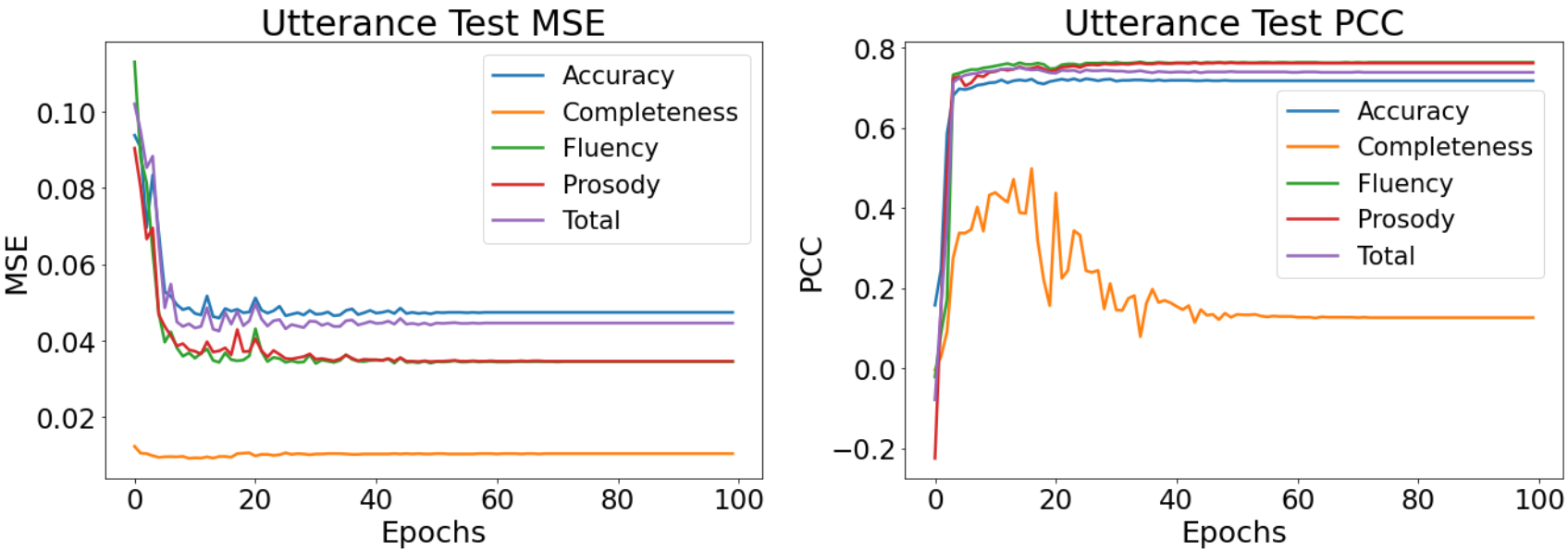}
    \caption{MSE and PCC score gap between different aspect-assessment tasks on the GOPT \cite{gong2022transformer} model. MSE is used as the loss function; however, for the \textit{Completeness} aspect, it indeed does not contribute to the training.}
    \label{fig2}
\end{figure}

Although the dataset promotes research on multi-aspect pronunciation assessment, the provided score labels are imbalanced, showing high-score-biased distributions, particularly in utterance \textit{Completeness} and word \textit{Stress} aspects (Figure~\ref{fig1}). Note that \textit{Completeness} actually has six labels but are not shown due to extremely huge samples of score 2. We attribute this imbalance to the notably inferior quality of scoring tasks of those two aspects in existing models \cite{gong2022transformer, chao20223m, 10095733}. Figure~\ref{fig2} shows that the mean squared error (MSE) loss, which is a generally used loss function for pronunciation assessment, is insufficient for scoring the high-score-biased \textit{Completeness} aspect. To overcome the limitations of the existing learning, we suggest SB loss, which re-weights the MSE loss and aids model training.

\subsection{Score-balanced loss}
To effectively re-weight the loss when training with imbalanced data, we introduce the SB factor. Our work is motivated by the class-balanced loss \cite{cui2019class}, which adds the weighting factor to the loss using the inverse of a demonstrated effective-number-of-samples term for an imbalanced visual classification task.

Unlike its application to the cross-entropy loss for classification, we extend the concept of the effective number of samples to the MSE loss for regression. Specifically, we re-scale the range of word and utterance aspect scores from 0-10 to 0-2 and regard scores at intervals of 0.2 as different classes, $\{0, 0.2, \cdots, 2\}$, defining two forms of score-balanced loss with different weighting factors: $SB_{num}$ and $SB_{rank}$. 

\subsubsection{Re-weighting by the number of samples: $SB_{num}$} According to a proven theoretical framework \cite{cui2019class}, the effective number of samples can be defined as $(1-\beta^n)/(1-\beta)$, where $n$ is the number of samples and $\beta$ is the hyperparameter defined with $(N-1)/N$. In contrast to class-balanced loss, which is weighted according to the number of samples $n$ for ground-truth class $y$ and assigned a fixed term irrespective of the direction of learning, we define $SB_{num}$ according to the number of samples of the \textit{predicted} score. Thus, the higher weights are assigned to the smaller number of samples of the predicted score class. By giving positive feedback for \textit{predicting} rare classes during training, the model is induced to predict more confidently regarding sparsely observed scores. Specifically, we consider predicted score $\hat{y}$ in the interval of $\left[s, s+0.2\right)$ as class $s$ and set the number of samples of score class $s$ ($n_{s}$) as $n_{\hat{y}}$. The $SB_{num}$ loss weight factor for the $m$-th aspect, $\alpha^m$, is defined as follows:
\begin{equation}
    \alpha^m = \begin{cases}
    {(1-\beta)}/{(1-\beta^{n_{\hat{y}}^m})} & , \text{if } n_{\hat{y}}^m \neq 0 \\
    1 & , \text{otherwise}
    \end{cases}
\end{equation}
where $n_{\hat{y}}^m$ denotes the number of samples for predicted score $\hat{y}$ on the $m$-th aspect. Then, $SB_{num}$ loss, where the MSE loss is multiplied by the SB factor can be defined as the following:
\begin{equation}
    SB_{num} = 
    \sum^{M}_{m=1}{\alpha^m \cdot (L_{MSE})^m}
\end{equation}
where $M$ is the number of all aspects at all phoneme, word, and utterance granularity levels; and $\alpha^m$ and $(L_{MSE})^m$ are the balancing factor and MSE loss of the $m$-th aspect, respectively.

\begin{table*}[t]
\caption{\label{tab1}
Experimental results of average MSE (only for phoneme-level) and PCC scores with standard deviation for five different runs. \textbf{Acc} is \textit{Accuracy} and \textbf{Comp} is \textit{Completeness}. \textbf{-baseline} denotes our implemented results of GOPT. \textbf{Bold} text denotes remarkably outperforming results over the baseline.}
\centering
\scalebox{
0.83}{
\begin{tabular}{l|cc|ccc|ccccc}
\toprule
& \multicolumn{2}{c|}{Phoneme Score} & \multicolumn{3}{c|}{Word Score (PCC)} & \multicolumn{5}{c}{Utterance Score (PCC)} \\
\hline
{Model} & Acc(MSE ↓) & Acc(PCC ↑) & Acc ↑ & Stress ↑ & Total ↑ & Acc ↑ & Comp ↑ & Fluency ↑ & Prosody ↑ & Total ↑ \\
\hline
\multirow{2}{*}{LSTM} & 0.089 & 0.587 & 0.511 & 0.297 & 0.524 & 0.717 & 0.123 & 0.741 & 0.744 & 0.743\\
 & \small{±0.002} & \small{±0.014} & \small{±0.014} & \small{±0.012} & \small{±0.011} & \small{±0.004} & \small{±0.143} & \small{±0.01} & \small{±0.006} & \small{±0.006} \\
\multirow{2}{*}{Gong et al.} & 0.085 & 0.612 & 0.533 & 0.291 & 0.549 & 0.714 & 0.155 & 0.753 & 0.760 & 0.742\\
 &  \small{±0.001} & \small{±0.003} & \small{±0.004} & \small{±0.030} & \small{±0.002} & \small{±0.004} & \small{±0.039} & \small{±0.008} & \small{±0.006} & \small{±0.005}\\
 \cline{2-11}
\multirow{2}{*}{\small{*}Gong et al.\small{-baseline}} & 0.086	& 0.609 & 0.526 & 0.272 & 0.543 & 0.717 & 0.134 & 0.755 & 0.756 & 0.739\\
& \small{±0.001} & \small{±0.003} & \small{±0.004} & \small{±0.048} & \small{±0.005} & \small{±0.006} & \small{±0.197} & \small{±0.006} & \small{±0.003} & \small{±0.004}\\
\hline
\multirow{2}{*}{\textbf{+$SB_{num}$}} & {0.086} & 0.605 & {0.531} & \textbf{0.386} & {0.547} & {0.722} & \textbf{0.427} & 0.750 & 0.752 & {0.747} \\
& \small{±0.001} & \small{±0.006} & \small{±0.005} & \small{±0.015} & \small{±0.005} & \small{±0.004} & \small{±0.101} & \small{±0.013} & \small{±0.007} & \small{±0.002}	\\
\cline{2-11}
\multirow{2}{*}{\textbf{+$SB_{rank}$}} & {0.086} & 0.605 & {0.529} & \textbf{0.341} & {0.544} & {0.717} & \textbf{0.335} & 0.750 & 0.745 & {0.743} \\
& \small{±0.001} & \small{±0.003} & \small{±0.004} & \small{±0.036} & \small{±0.002} & \small{±0.007} & \small{±0.235} & \small{±0.005} & \small{±0.015} & \small{±0.007}	\\
\bottomrule
\end{tabular}}
\end{table*}

\subsubsection{Re-weighting by the ranking: $SB_{rank}$} Considering that the variance of the number of samples is large depending on the classes, we further design $SB_{rank}$ to assign weights according to ranks rather than the actual sample number per class. As there are only 11 classes (from 0 to 2), the rank gap between classes is not significant compared with up to 2500 sample gaps; therefore, replacing balancing criteria with rank enables smooth learning. 

Specifically, we rank the number of samples of each class in descending order and use the inverse rank as an SB term; therefore, the majority label is weighted less. For the same values, the average-rank policy was applied (e.g., rank 1.5 is assigned to two classes of the same smallest number of samples). For the obtained ranking of the number of samples of the predicted class $\hat{y}$, normalization is applied. The weighting factor of $SB_{rank}$ loss when predicting the $m$-th aspect is defined as
\begin{equation}
\gamma^m = (1/r_{\hat{y}}^m)
\end{equation}
where $r_{\hat{y}}^m$ is the normalized ranking of predicted class size on the $m$-th aspect. Then, the total $SB_{rank}$ loss across all M aspects is represented as follows:
\begin{equation}
    SB_{rank} = 
    \sum^{M}_{m=1}{\gamma^m \cdot (L_{MSE})^m}
\end{equation}

\section{Experiments}
We evaluate our SB loss on the publicly available speechocean762 \cite{zhang2021speechocean762} dataset, which has imbalanced labels and is well-built for the multi-aspect pronunciation assessment task. We use the public GOPT\footnote{\url{https://github.com/YuanGongND/gopt}} model \cite{gong2022transformer}, which predicts multiple aspect scores in parallel. GOPT is based on a transformer \cite{vaswani2017attention} architecture and uses the goodness of pronunciation (GOP) features. In detail, with the audio and corresponding canonical transcription input, the acoustic model outputs frame-level phonetic posterior probabilities, which are then transformed into GOP features with 84 dimensions. Then, GOP features projected to 24 dimensions, canonical phoneme embedding, and positional embedding are added and input to the three-layer transformer encoder with 24 embedding dimensions. 

We follow the same settings described for GOPT experiments, except for the loss function and graphics processing unit (GPU). In detail, the Adam optimizer, an initial 1e-3 learning rate, 25 batch size, and 100 epochs are set for training. All models have 26.577k parameters. The automatic speech recognition model\footnote{\url{https://kaldi-asr.org/models/m13}} trained with LibriSpeech \cite{panayotov2015librispeech} 960-hour data is used. Training and test sets, each comprising 2500 utterances, are used. NVIDIA RTX A5000 GPU is used for all experiments. Five different experiments are conducted by different random seeds, and their mean and standard deviation are reported. PCC is used for the evaluation metric; additionally, MSE is used for phoneme-level accuracy. We set hyperparameter $\beta$ in $S_{num}$ as 0.9 after conducting experiments with diverse values.

\begin{figure}[t]
    \centering \includegraphics[width=8cm]{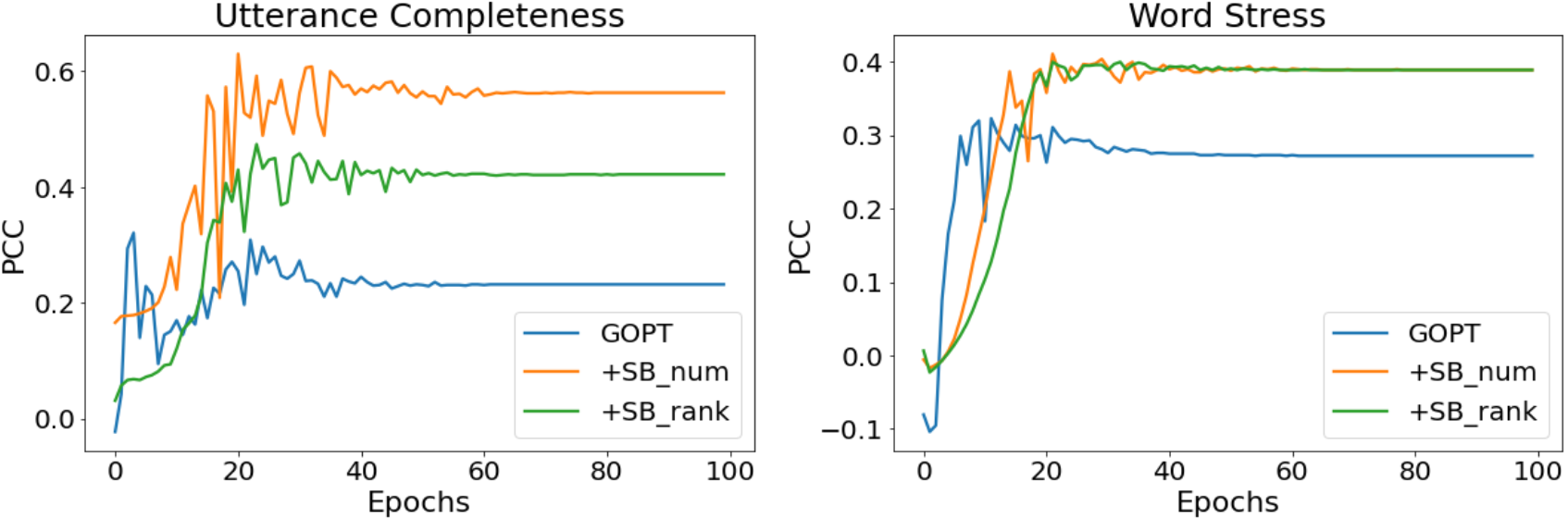}
    \caption{Test PCC scores of utterance \textit{Completeness} and word \textit{Stress} aspect according to training epochs.}
    \label{fig3}
\end{figure}

\section{Results and discussion}
To compare the results in the same computational environment, we re-implement the GOPT model and use for comparison. Table~\ref{tab1} reveals that both models trained with $SB_{num}$ and $SB_{rank}$ loss exhibit exceptional improvements regarding aspects with extremely unbalanced score distributions, indicating that our approach effectively tackles the negative impact of imbalanced data. Specifically, after training with $SB_{num}$ loss, absolute improvements of 11.4\% and 29.3\% are observed for the word \textit{Stress} and utterance \textit{Completeness}, respectively. This further reduces the performance gap between different tasks in a multi-aspect learning setting, as the difference between the most superior (\textit{Prosody}) and inferior (\textit{Completeness}) aspect is reduced from 5 to 1.7 times. Meanwhile, there are no significant PCC score differences over the baseline in aspects where samples are somewhat present even in minority score classes. Similar trends are observed for the model trained with $SB_{rank}$ loss, which shows absolute improvements of 6.9\% and 20.1\% in \textit{Stress} and \textit{Completeness}, respectively. Overall evaluation results for aspects indicate that specifying the number of samples in each scoring class allows more precise re-weighting than abstracting them with the rank. The changes in the test PCC score by the learning process (Figure~\ref{fig3}) reveal that $SB_{num}$ and $SB_{rank}$ converge to a specific value at the latter and higher point than the baseline, implying that the training validity continued longer.

\begin{table*}[t]
\caption{\label{tab2}
Comparison results between $SB_{num}$ with the sample number of predicted $\hat{y}$ ($SB_{num}$) and $SB_{num}$ with the sample number of ground-truth $y$ ($SB_{num}$ (no pred)). \textbf{Bold} text denotes the particularly improved results.}
\centering
\scalebox{
0.8}{
\begin{tabular}{l|cc|ccc|ccccc}
\toprule
& \multicolumn{2}{c|}{Phoneme Score} & \multicolumn{3}{c|}{Word Score (PCC)} & \multicolumn{5}{c}{Utterance Score (PCC)} \\
\hline
{Model} & Acc(MSE ↓) & Acc(PCC ↑) & Acc ↑ & Stress ↑ & Total ↑ & Acc ↑ & Comp ↑ & Fluency ↑ & Prosody ↑ & Total ↑ \\
\hline
\multirow{2}{*}{+$SB_{num}$ \small{(no pred)}} & {0.085} & {0.611} & {0.534} & {0.293} & {0.549} & {0.717} & {0.218} & {0.760} & {0.758} & {0.745} \\
& \small{±0.000} & \small{±0.002}  & \small{±0.006} & \small{±0.031} & \small{±0.004} & \small{±0.006} & \small{±0.097} & \small{±0.007} & \small{±0.011} & \small{±0.006} \\
\hline
\multirow{2}{*}{{+$SB_{num}$}} & {0.086} & 0.605 & {0.531} & \textbf{0.386} & {0.547} & {0.722} & \textbf{0.427} & 0.750 & 0.752 & {0.747} \\
& \small{±0.001} & \small{±0.006} & \small{±0.005} & \small{±0.015} & \small{±0.005} & \small{±0.004} & \small{±0.101} & \small{±0.013} & \small{±0.007} & \small{±0.002}	\\

\bottomrule
\end{tabular}}
\end{table*}

\subsection{Ablation study}
Apart from the proposed $SB_{num}$ loss, which is based on the predicted value, we additionally experiment with SB loss using the ground-truth label of each sample. Instead of the number of samples in the class of the predicted $\hat{y}$, that of ground-truth $y$ is used; therefore, the weighting factor for the $m$-th aspect, $\alpha^m$, is defined as $(1-\beta)/(1-\beta^{n_{y}^m})$. Note that cases are not needed here, because all samples satisfy $n_{y}^m\geq1$. 

Table~\ref{tab2} shows almost similar PCC scores across most aspects, except for \textit{Stress} and \textit{Completeness}. For these two aspects, our prediction-based approach achieved higher performance than the label-based application. This implies that dynamically re-weighting the loss according to predicted values for each sample during training is more effective than assigning fixed weights from the beginning.

\subsection{Effects of different $\beta$ values}
\begin{figure}[t]
    \centering \includegraphics[width=7.1cm]{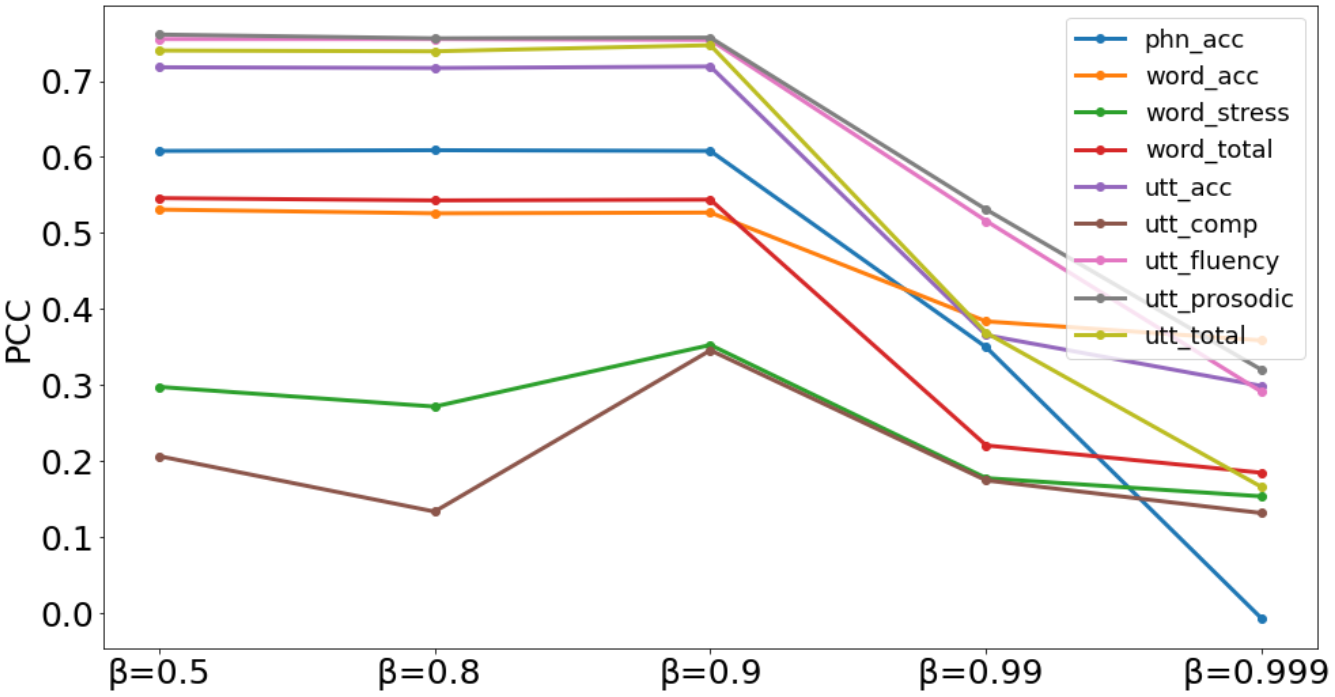}
    \caption{Change in PCC scores of the model with $SB_{num}$ according to change in hyperparameter $\beta$ value for each aspect.}
    \label{fig4}
\end{figure}

We observe the effects of different hyperparameter $\beta$ values when training with $SB_{num}$ loss. Figure~\ref{fig4} shows the best performance at $\beta=0.9$, and thereafter, the PCC values sharply decrease with an increase in $\beta$. This can be explained by the variations on the scale of weight factors. When the value of $\beta$ is greater, the scale of the weighting factor $\alpha$ becomes smaller, and vice versa (Table~\ref{tab3}). Thus, increasing the $\beta$ value hinders training in the case of the proposed SB loss. 

\begin{table}[t]
\caption{\label{tab3}
Averaged $(1-\beta)/(1-\beta^{n_{y}})$ values across 11 classes for different $\beta$ in the case of the phoneme-level \textit{Accuracy} aspect.}
\centering
\scalebox{
0.78}{
\begin{tabular}{l|c|c|c|c|c}
\toprule
\diagbox[width=12em]{}{$\beta$} & 0.5 &	0.8 &	0.9 &	0.99 &	0.999  \\
\midrule
Avg $(1-\beta)/(1-\beta^{n_{y}})$ & 0.542 & 	0.267 & 	0.175 & 	0.095 & 	0.088 \\
\bottomrule
\end{tabular}}
\end{table}

\subsection{Qualitative evaluation}
\begin{figure}[t]
    \centering \includegraphics[width=8cm]{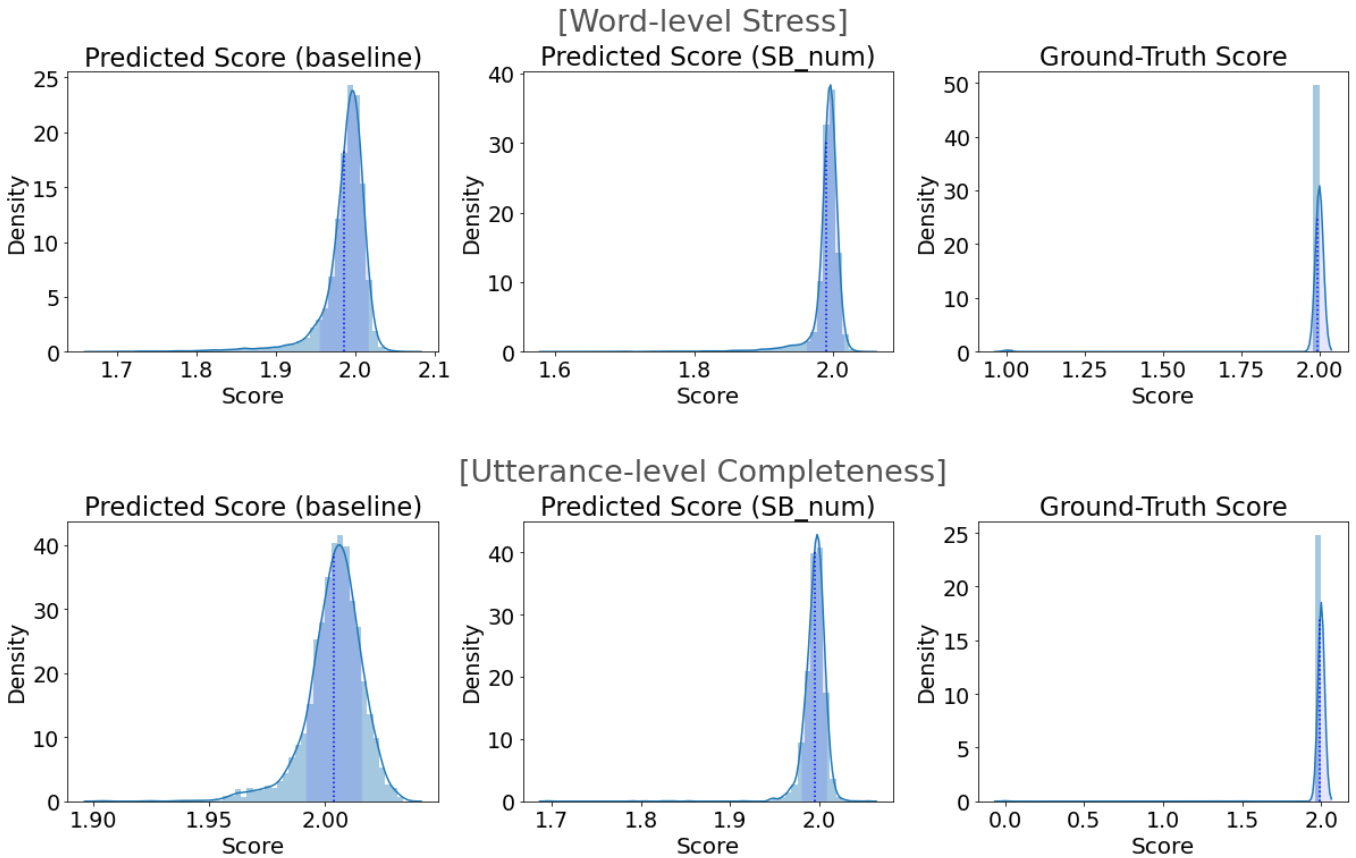}
    \caption{Comparison of predicted score distributions (of baseline and our method) and test set ground-truth scores; vertical lines represent mean, darker area shows standard deviations.}
    \label{fig5}
\end{figure}

To examine the actual mechanism behind our model aiding in pronunciation evaluation, we conduct a qualitative analysis focusing on \textit{Completeness} and \textit{Stress}. For each aspect, we compare the predicted score distribution of the baseline model and that of the $SB_{num}$-loss-trained model with the actual ground-truth score on the test set. Figure~\ref{fig5} reveals that the model with our approach predicts a wider range of scores while approximating closer to the predicted score distribution, which is densely distributed around the score of 2. This comparison indicates that our strategy not only helps predict minority samples but also supports accurate predictions for majority samples. 

In particular, for the word-level \textit{Stress}, the actual target scores are distributed in the range 1-2. The lowest score predicted by the baseline is 1.67, while that by our method is 1.58, which is closer to the actual score of 1. For utterance-level \textit{Completeness}, the lowest predicted score by the baseline is 1.90, while that with ours is 1.69, both of which are the samples of the ground-truth 0 score. Samples with \textit{Completeness} score of 0 do not exist in the training set, thus our closer prediction implies the assistance on missing data at continuous target values.

\section{Conclusions}
In this paper, we propose a simple yet effective score-balanced loss function for multi-aspect pronunciation assessment. Assuming continuous scores as the categorical classes, we design two forms of SB loss, assigning high costs when the model predicts minor classes. Both re-weighting by exploiting the number of samples in the predicted value and by the ranks of sample size effectively balance the biased training. The experimental results showing remarkable improvement on the highly imbalanced aspects prove that SB loss overcomes the limitation of unevenly distributed datasets.

\noindent \\ \textbf{Acknowledgements:}
This work was partly supported by Institute of Information \& communications Technology Planning \& Evaluation (IITP) grant funded by the Korea government (MSIT) (No.2019-0-01906, Artificial Intelligence Graduate School Program (POSTECH)) and MSIT (Ministry of Science and ICT), Korea, under the ITRC (Information Technology Research Center) support program (IITP-2023-2020-0-01789) supervised by the IITP.

\bibliographystyle{IEEEtran}
\bibliography{main}

\begin{thebibliography}{10}
\providecommand{\url}[1]{#1}
\csname url@samestyle\endcsname
\providecommand{\newblock}{\relax}
\providecommand{\bibinfo}[2]{#2}
\providecommand{\BIBentrySTDinterwordspacing}{\spaceskip=0pt\relax}
\providecommand{\BIBentryALTinterwordstretchfactor}{4}
\providecommand{\BIBentryALTinterwordspacing}{\spaceskip=\fontdimen2\font plus
\BIBentryALTinterwordstretchfactor\fontdimen3\font minus
  \fontdimen4\font\relax}
\providecommand{\BIBforeignlanguage}[2]{{%
\expandafter\ifx\csname l@#1\endcsname\relax
\typeout{** WARNING: IEEEtran.bst: No hyphenation pattern has been}%
\typeout{** loaded for the language `#1'. Using the pattern for}%
\typeout{** the default language instead.}%
\else
\language=\csname l@#1\endcsname
\fi
#2}}
\providecommand{\BIBdecl}{\relax}
\BIBdecl

\bibitem{eskenazi2009overview}
M.~Eskenazi, ``An overview of spoken language technology for education,''
  \emph{Speech Communication}, vol.~51, no.~10, pp. 832--844, 2009.

\bibitem{franco1997automatic}
H.~Franco, L.~Neumeyer, Y.~Kim, and O.~Ronen, ``Automatic pronunciation scoring
  for language instruction,'' in \emph{1997 IEEE international conference on
  acoustics, speech, and signal processing}, vol.~2.\hskip 1em plus 0.5em minus
  0.4em\relax IEEE, 1997, pp. 1471--1474.

\bibitem{witt2000phone}
S.~M. Witt and S.~J. Young, ``Phone-level pronunciation scoring and assessment
  for interactive language learning,'' \emph{Speech communication}, vol.~30,
  no. 2-3, pp. 95--108, 2000.

\bibitem{shi2020context}
J.~Shi, N.~Huo, and Q.~Jin, ``Context-aware goodness of pronunciation for
  computer-assisted pronunciation training,'' \emph{arXiv preprint
  arXiv:2008.08647}, 2020.

\bibitem{sancinetti2022transfer}
M.~Sancinetti, J.~Vidal, C.~Bonomi, and L.~Ferrer, ``A transfer learning
  approach for pronunciation scoring,'' in \emph{ICASSP 2022-2022 IEEE
  International Conference on Acoustics, Speech and Signal Processing
  (ICASSP)}.\hskip 1em plus 0.5em minus 0.4em\relax IEEE, 2022, pp. 6812--6816.

\bibitem{tepperman2005automatic}
J.~Tepperman and S.~Narayanan, ``Automatic syllable stress detection using
  prosodic features for pronunciation evaluation of language learners,'' in
  \emph{Proceedings.(ICASSP'05). IEEE International Conference on Acoustics,
  Speech, and Signal Processing, 2005.}, vol.~1.\hskip 1em plus 0.5em minus
  0.4em\relax IEEE, 2005, pp. I--937.

\bibitem{cucchiarini2000quantitative}
C.~Cucchiarini, H.~Strik, and L.~Boves, ``Quantitative assessment of second
  language learners’ fluency by means of automatic speech recognition
  technology,'' \emph{The Journal of the Acoustical Society of America}, vol.
  107, no.~2, pp. 989--999, 2000.

\bibitem{arias2010automatic}
J.~P. Arias, N.~B. Yoma, and H.~Vivanco, ``Automatic intonation assessment for
  computer aided language learning,'' \emph{Speech communication}, vol.~52,
  no.~3, pp. 254--267, 2010.

\bibitem{li2017intonation}
K.~Li, X.~Wu, and H.~Meng, ``Intonation classification for l2 english speech
  using multi-distribution deep neural networks,'' \emph{Computer Speech \&
  Language}, vol.~43, pp. 18--33, 2017.

\bibitem{gong2022transformer}
Y.~Gong, Z.~Chen, I.-H. Chu, P.~Chang, and J.~Glass, ``Transformer-based
  multi-aspect multi-granularity non-native english speaker pronunciation
  assessment,'' in \emph{ICASSP 2022-2022 IEEE International Conference on
  Acoustics, Speech and Signal Processing (ICASSP)}.\hskip 1em plus 0.5em minus
  0.4em\relax IEEE, 2022, pp. 7262--7266.

\bibitem{chao20223m}
F.-A. Chao, T.-H. Lo, T.-I. Wu, Y.-T. Sung, and B.~Chen, ``3m: An effective
  multi-view, multi-granularity, and multi-aspect modeling approach to english
  pronunciation assessment,'' in \emph{2022 Asia-Pacific Signal and Information
  Processing Association Annual Summit and Conference (APSIPA ASC)}.\hskip 1em
  plus 0.5em minus 0.4em\relax IEEE, 2022, pp. 575--582.

\bibitem{10095733}
H.~Do, Y.~Kim, and G.~G. Lee, ``Hierarchical pronunciation assessment with
  multi-aspect attention,'' in \emph{ICASSP 2023 - 2023 IEEE International
  Conference on Acoustics, Speech and Signal Processing (ICASSP)}, 2023, pp.
  1--5.

\bibitem{yang2014machine}
X.~Yang, A.~Loukina, and K.~Evanini, ``Machine learning approaches to improving
  pronunciation error detection on an imbalanced corpus,'' in \emph{2014 IEEE
  Spoken Language Technology Workshop (SLT)}.\hskip 1em plus 0.5em minus
  0.4em\relax IEEE, 2014, pp. 300--305.

\bibitem{wang2017learning}
Y.-X. Wang, D.~Ramanan, and M.~Hebert, ``Learning to model the tail,''
  \emph{Advances in neural information processing systems}, vol.~30, 2017.

\bibitem{cui2019class}
Y.~Cui, M.~Jia, T.-Y. Lin, Y.~Song, and S.~Belongie, ``Class-balanced loss
  based on effective number of samples,'' in \emph{Proceedings of the IEEE/CVF
  conference on computer vision and pattern recognition}, 2019, pp. 9268--9277.

\bibitem{zhang2021speechocean762}
J.~Zhang, Z.~Zhang, Y.~Wang, Z.~Yan, Q.~Song, Y.~Huang, K.~Li, D.~Povey, and
  Y.~Wang, ``speechocean762: An open-source non-native english speech corpus
  for pronunciation assessment,'' \emph{arXiv preprint arXiv:2104.01378}, 2021.

\bibitem{he2009learning}
H.~He and E.~A. Garcia, ``Learning from imbalanced data,'' \emph{IEEE
  Transactions on knowledge and data engineering}, vol.~21, no.~9, pp.
  1263--1284, 2009.

\bibitem{japkowicz2002class}
N.~Japkowicz and S.~Stephen, ``The class imbalance problem: A systematic
  study,'' \emph{Intelligent data analysis}, vol.~6, no.~5, pp. 429--449, 2002.

\bibitem{padurariu2019dealing}
C.~Padurariu and M.~E. Breaban, ``Dealing with data imbalance in text
  classification,'' \emph{Procedia Computer Science}, vol. 159, pp. 736--745,
  2019.

\bibitem{chawla2002smote}
N.~V. Chawla, K.~W. Bowyer, L.~O. Hall, and W.~P. Kegelmeyer, ``Smote:
  synthetic minority over-sampling technique,'' \emph{Journal of artificial
  intelligence research}, vol.~16, pp. 321--357, 2002.

\bibitem{estabrooks2004multiple}
A.~Estabrooks, T.~Jo, and N.~Japkowicz, ``A multiple resampling method for
  learning from imbalanced data sets,'' \emph{Computational intelligence},
  vol.~20, no.~1, pp. 18--36, 2004.

\bibitem{ren2022balanced}
J.~Ren, M.~Zhang, C.~Yu, and Z.~Liu, ``Balanced mse for imbalanced visual
  regression,'' in \emph{Proceedings of the IEEE/CVF Conference on Computer
  Vision and Pattern Recognition}, 2022, pp. 7926--7935.

\bibitem{yang2021delving}
Y.~Yang, K.~Zha, Y.~Chen, H.~Wang, and D.~Katabi, ``Delving into deep
  imbalanced regression,'' in \emph{International Conference on Machine
  Learning}.\hskip 1em plus 0.5em minus 0.4em\relax PMLR, 2021, pp.
  11\,842--11\,851.

\bibitem{basuki2018use}
Y.~Basuki, ``The use of drilling method in teaching phonetic transcription and
  word stress of pronunciation class,'' \emph{Karya Ilmiah Dosen}, vol.~1,
  no.~1, 2018.

\bibitem{vaswani2017attention}
A.~Vaswani, N.~Shazeer, N.~Parmar, J.~Uszkoreit, L.~Jones, A.~N. Gomez,
  {\L}.~Kaiser, and I.~Polosukhin, ``Attention is all you need,''
  \emph{Advances in neural information processing systems}, vol.~30, 2017.

\bibitem{panayotov2015librispeech}
V.~Panayotov, G.~Chen, D.~Povey, and S.~Khudanpur, ``Librispeech: an asr corpus
  based on public domain audio books,'' in \emph{2015 IEEE international
  conference on acoustics, speech and signal processing (ICASSP)}.\hskip 1em
  plus 0.5em minus 0.4em\relax IEEE, 2015, pp. 5206--5210.

\end{thebibliography}

\end{document}